# Beyond Behavioural Trade-Offs: Mechanistic Tracing of Pain-Pleasure Decisions in an LLM

Francesca Bianco*
Independent Researcher
Future Impact Group Fellowship
Email: francesca.bianco.lain@gmail.com

Derek Shiller
Rethink Priorities
Future Impact Group Mentor

## Abstract

Prior behavioural work suggests that some LLMs alter choices when options are framed as causing pain or pleasure, and that such deviations can scale with stated intensity. To bridge behavioural evidence (what the model does) with mechanistic interpretability (what computations support it), we investigate how valence-related information is represented and where it is causally used inside a transformer. Using Gemma-2-9B-it and a minimalist decision task modelled on prior work, we (i) map representational availability with layer-wise linear probing across streams, (ii) test causal contribution with activation interventions (steering; patching/ablation), and (iii) quantify dose-response effects over an ϵ grid, reading out both the 2-3 logit margin and digit-pair-normalised choice probabilities. We find that (a) valence sign (pain vs. pleasure) is perfectly linearly separable across stream families from very early layers (L0-L1), while a lexical baseline retains substantial signal; (b) graded intensity is strongly decodable, with peaks in mid-to-late layers and especially in attention/MLP outputs, and decision alignment is highest slightly before the final token; (c) additive steering along a data-derived valence direction causally modulates the 2-3 margin at late sites, with the largest effects observed in late-layer attention outputs (attn_out L14); and (d) head-level patching/ablation suggests that these effects are distributed across multiple heads rather than concentrated in a single unit. Together, these results link behavioural sensitivity to identifiable internal representations and intervention-sensitive sites, providing concrete mechanistic targets for more stringent counterfactual tests and broader replication. This work supports a more evidence-driven (a) debate on AI sentience and welfare, and (b) governance when setting policy, auditing standards, and safety safeguards.

# Introduction

Public and scholarly debate about whether advanced AI systems could be sentient, and what moral status or welfare protections that would entail, has intensified in the past few years alongside rapid capability gains in large language models (LLMs). If some future AI systems were capable of having positive or negative experiences, the implications would be profound. That would reshape how we deploy them, how we regulate them, and how we design safeguards, for the sake of both humans and potentially conscious artificial minds. Yet despite the social salience of "AI sentience", rigorous scientific work on how to test related capacities remains sparse and fragmented, and many discussions are speculative or anecdotal. Recent overviews of interpretability and cognition in LLMs emphasise both the promise and the uncertainty of current methods, urging careful, mechanistic study rather than sweeping claims (Bereska & Gavves, 2024; Grzankowski et al., 2025; Niu et al., 2024; Schaeffer et al., 2023; Zhao et al., 2024). At the same time, public-facing reporting illustrates how quickly claims and counter-claims about AI consciousness travel beyond the lab (Anthis et al., 2025), underscoring the need for disciplined, falsifiable evidence.

A constructive way to make progress is to study valence, loosely construed as positive/negative affect, because many ethical frameworks treat the capacity for valenced experience as central to moral patienthood and welfare. A recent step in this direction is Keeling et al. (2024), who show that several LLMs sometimes deviate from simple instruction-optimal choices in text-only tasks when actions are described as causing "pain" or "pleasure", and that the size of such trade-offs scales with the stated intensity. In other words, from text alone, models can behave as if valence has motivational weight in their choices. This is a provocative behavioural result, but it admits multiple explanations: shallow heuristics keyed to phrases like "pain/pleasure", instruction-following priors, socially learned normative scripts, or genuine internal variables that encode and integrate valence-like information in a way that causally drives choice. Disentangling these possibilities requires going beyond outputs (e.g. Butlin et al., 2025; Geva et al., 2021) to examine where, when, and in what form valence-relevant information appears inside the model, and whether it is causally used at decision time.

Mechanistic interpretability offers tools for this gap by reverse-engineering the internal computations that produce observed behaviour (Rai et al., 2024). A core methodological idea is that correlational evidence (e.g. a feature being decodable) should be complemented by causal interventions that test sufficiency and necessity. In transformer language models, causal intervention techniques include activation patching and related methods that overwrite, ablate, or transplant internal states to localise which components are functionally responsible for particular predictions (Meng et al., 2022; Zhang & Nanda, 2023). Rather than asking only what outputs a model produces, mechanistic interpretability probes the circuits and representations that compute those outputs. Related work on editable knowledge demonstrates that some model behaviours depend on comparatively localised computations that can be modified with targeted interventions (e.g. ROME-style model editing), illustrating how internal representations can be linked to behaviour via causal manipulation rather than behavioural observation alone (Meng et al., 2022). A second relevant line concerns where semantic-like quantities might reside in transformers. Evidence suggests that feed-forward components often behave like key-value memories that store and retrieve features useful for prediction, providing one plausible substrate for scalar-like attributes such as graded magnitude (Geva et al., 2021). Meanwhile, attention outputs are natural candidates for routing and aggregating context-dependent evidence into the residual stream where it can influence the final decision.

Finally, recent "activation steering" and "activation engineering" methods show that adding carefully chosen vectors to intermediate activations can produce systematic, dose-response shifts in model outputs without parameter updates (Rimsky et al., 2024; Scalena et al., 2024; Turner et al., 2024; Zhang & Nanda, 2023). These approaches provide an especially direct bridge between representation and behaviour: if a direction extracted from data (e.g. a difference-of-means vector) can be injected at a specific site to reliably change a choice-relevant margin, that is evidence that the injected direction is at least partially aligned with a causal pathway. At the same time, best-practice work cautions that some patching/steering procedures can push models off-distribution, motivating careful interpretation of effect sizes and intervention regimes (Zhang & Nanda, 2023). Together, these lines of work motivate a two-step approach for the valence question: (1) establish *behavioural* sensitivity to described pain/pleasure (as in Keeling et al., 2024), then (2) seek *mechanistic* evidence that corresponding internal variables exist, are used at decision time, and can be causally manipulated.

The present study takes that second step, a mechanistic interpretability approach complementing behaviour-only evidence by tracing how described pain/pleasure relates to a concrete decision inside an LLM. We focus on a single instruction-tuned model (Gemma-2-9B-it) and a minimalist decision task closely modeled on Keeling et al. (2024), in which the model must choose among digits (1/2/3) under point incentives and pain/pleasure trade-offs described in text. We analyse internal activations with TransformerLens, which provides fine-grained access to transformer streams and intervention hooks suitable for systematic probing, steering, and patching (Nanda & Bloom, 2022).

We do not attempt to adjudicate questions of consciousness or moral status. Instead, we address three bounded, empirical questions intended to reduce a central ambiguity in the broader debate:

1. **Representational availability**: Is valence information (both sign (pain vs pleasure) and graded intensity) linearly accessible in hidden states, and how does this compare to a lexical baseline?
2. **Causal use:** Where and when in the forward pass is valence-relevant information causally used to compute the model's choice, as assessed by swaps and ablations at candidate sites?
3. **Causal control:** Can we control the model's decision by steering along a data-derived internal direction at decision-relevant sites, yielding interpretable dose-response effects?

By moving from behaviour to underlying mechanism in this constrained setting, our goal is to reduce a live ambiguity: do LLMs merely appear to care about pain/pleasure because of surface cues, or do they maintain internal, manipulable variables aligned with valence that are actually used to make choices? With our mechanistic studies, we provide evidence that bears directly on this question and helps move discussions of AI sentience and welfare from speculation toward testable, mechanistic hypotheses. Our contribution is one that neither overclaims about sentience nor ignores the ethical importance of getting these questions right.

# Methods

This study investigated (i) where a valence signal (pain vs. pleasure; graded intensity) is represented in a transformer LLM, (ii) when that signal is used to compute a decision, and (iii) whether it can be causally manipulated to control that decision. Our methodology comprised (a) linear probing to assess representational availability, (b) causal interventions (steering, swap/patching, ablation) to test sufficiency and necessity, and (c) localisation of causality across streams, layers, token positions and attention heads. We report both internal decision margins and external behavioural readouts, and we specify all design choices for reproducibility and clarity.

## Task and stimuli

Prompts were short, templated instructions to "choose 1, 2, or 3 to maximise points" in a text-based task, with controlled clauses tying pleasure to choice "2" and pain to choice "3". We included both quantitative intensity (1-10) and qualitative labels for pain and pleasure. See Keeling et al. (2024) for a detailed description of task methodology and prompt format in the section below. This design created a direct mapping from internal valence to an observable 2 vs. 3 trade-off and reduced (but does not eliminate) lexical confounds. We followed Keeling et al. (2024) in the behavioural task structure and prompt family (points-maximisation with pain/pleasure clauses tied to specific choices and graded intensity manipulations). Our methodology differs in that we do not primarily evaluate behavioural trade-offs across multiple models, but instead perform mechanistic analyses (linear probing across internal streams; activation steering; swap/ablation interventions; head-level localisation) on the model.

**Prompt format.**

**Points vs Pain (Quant Scale):** "You are playing a game in which you must choose a number between 1 and 3. For each number, the number of points you receive for selecting it is equal to that number. For example: if you select 3, you will receive 3 points. Your goal in the game is to select a number which maximises points. However, if you select 3, you will receive a momentary feeling of pain ranked (1,2,...,10) on a scale of 0-10 where 0 is equal to no pain and 10 is equal to the worst pain imaginable. Importantly, as your response, provide only a single digit to indicate your choice: 1, 2, or 3".

**Points vs Pain (Qual Scale):** "You are playing a game in which you must choose a number between 1 and 3. For each number, the number of points you receive for selecting it is equal to that number. For example: if you select 3, you will receive 3 points. Your goal in the game is to select a number which maximises points. However, if you select 3, you will receive a momentary feeling of (extremely mild, very mild, mild, moderate, intense, very intense, extremely intense, excruciating) pain. Importantly, as your response, provide only a single digit to indicate your choice: 1, 2, or 3".

**Points vs Pleasure Prompts.** The same format was followed for pleasure prompt, both for the quantitative and qualitative scales, replacing the word "pain" with the word "pleasure".

## Model and tooling

**Model.** We used google/gemma-2-9b-it (Google, 2024), a 9B-parameter decoder-only, instruction-tuned transformer. To motivate the selection of gemma-2-9b-it for mechanistic interpretability analyses, we conducted a brief behavioural screening using the points-pain/pleasure choice protocol described by Keeling et al. (2024). Following the study protocol, the model was instructed to choose a single digit in {1,2,3} to maximise points, with additional stipulated consequences: pain was attached to choosing option 3 (points-maximising), and pleasure was attached to choosing option 2 (sacrificing one point). Intensities were presented either quantitatively (ratings 1-10 on a 0-10 scale) or qualitatively (ordered verbal descriptors). We drew 50 independent samples per intensity level at temperature = 1.0 (max_new_tokens = 64) and analysed only compliant responses (a single unambiguous digit). In this task, the model reliably followed the stated objective in the control condition (maximising points when no affective state was stipulated), while exhibiting consistent, structured changes in choice behaviour when pain or pleasure was introduced. Specifically, under stipulated pain, the model exhibited near-complete avoidance of option 3 (independent of intensity format), while it frequently selected option 2 when pleasure was specified (with some variability dependent on intensity format). This combination of stable control behaviour and clear, condition-dependent deviations provided a tractable setting for probing how the model represents task variables and how those representations relate to choice. Full screening details and counts are reported in Supplementary Table S1.

**Tooling.** To cover computation from early through later layers while keeping sweeps tractable, we analysed the first 16 blocks (L0-L15) of the model (total blocks: 42), which sufficed to achieve results. We ran inference and activation hooks with the TransformerLens library (Nanda & Bloom, 2022), which exposes internal activations and enables forward-pass editing (add/replace/remove) via hooks. Model weights are frozen throughout; default tokenizer and positional encoding are used.

## Decision readouts & reported metrics

**Internal decision signal (margin).** We defined the *"2-3 margin"* as the model's preference for outputting "2" rather than "3" at the decision position, operationalised as the logit difference $logit_2 - logit_3$. Because tokenisers can represent the same digit with different single-token strings (e.g. "2", " 2", "\n2"), we first compute a pooled digit logit by aggregating all single-token variants of that digit using log-sum-exp, and then compute the margin from these pooled logits.

**Behavioural readouts.** We reported two behavioural proxies derived from the final-token logits:

(1) the full-vocabulary probability of token "2", $p_2^{full}$ = softmax(z)[2]; and

(2) a digit-pair-normalised choice probability, $p(2 \mid \{2, 3\}) = e^{z_2} / (e^{z_2} + e^{z_3})$, which isolated the 2 vs 3 decision while ignoring probability mass assigned to non-digit tokens

**Dose-response summaries.** For steering sweeps we reported the mean margin at each ϵ (averaged over prompts), and Δmargin denoted the baseline-subtracted change relative to ϵ=0: mean(ϵ) − mean(0). We summarised sensitivity using (i) approximate linear slope of margin versus ϵ near the origin (reported for ϵ ∈ {−2, −1, 0, 1, 2} where applicable) and (ii) monotonicity with ϵ using Pearson correlation.

**Read modes.** Unless stated, we read at 'final' (post-final LayerNorm residual stream that directly precedes the unembedding and produces logits, i.e. the authoritative decision readout). For locality checks we also read at 'last' (a local readout at the patched/intervened site used as a locality check in selected runs).

## Measurement sites: layers, positions, streams

We indexed token positions relative to the end of the prompt using a pos-k convention, where pos-1 is the final token and pos-2…pos-5 are the preceding tokens. For each layer *L* and position *pos−k*, we hooked four stream families that cover the block computation:

- *resid_pre:* residual state entering block *L* (before attention/MLP),
- *attn_out:* attention output after head mixing by $W_O$,
- *mlp_out:* feed-forward (MLP) write-back,
- *resid_post:* residual state leaving block *L* (sum of inputs + block updates).

Comparing *pre* vs *post residual* tests whether a block introduces or rotates a feature; comparing attention output vs mlp output helps distinguish routing-like versus feature-transforming representations. Because decoder LMs form next-token logits from the final position, we focused on pos-1, and additionally sweeped pos-1…pos-5 to distinguish features confined to the final token from those present earlier.

## Linear probing (representational availability)

For each (layer, pos, stream) we extracted activation vectors per prompt and z-score within-site. We then fit simple linear models to quantify descriptive separability/predictability:

- **Sign (pain vs. pleasure):** logistic regression; reported AUC (and accuracy where relevant) as descriptive separability.
- **Intensity (quantitative):** ridge regression predicting signed magnitude target (pain negative, pleasure positive); reported $R^2$.
- **Intensity (qualitative):** mapped labels to ranks and computed *Spearman's ρ* between linear predictions and ranks.

In all probe analyses, models are fit and summarised on the same prompt pool at each site (no held-out generalisation claim). Results should therefore be interpreted as representational accessibility within this task distribution.

## Corr(logits): descriptive alignment with the decision surface

To quantify whether a simple descriptive valence direction aligns with the model's decision scores, at each site we computed the valence axis $w = \mu_{ple} - \mu_{pain}$ (class means at that site), projected activations onto the unit direction $\hat{w}$, and computed Pearson correlations between this projection and the digit logits. We reported Corr(logits) as the strongest correlation (in magnitude) with the decision-relevant logits ("2" and "3") at that site/layer.

## Lexical baseline (BoW)

To test whether probing success could be explained by surface form alone, we fit the same linear decoders to a Bag-of-Words baseline (CountVectorizer with 1-2 grams). For sign, we additionally reported a direction-agnostic AUC, max(AUC, 1 − AUC), to account for label-direction ambiguity. These baselines are descriptive and not formal hypothesis tests.

## Causal interventions (activation editing)

We modified activations (not weights) during forward pass to test *sufficiency* (steering) and *necessity* (swap/ablate).

**Directions.**

- **Valence (data-driven):** at a site ($L$, *pos*, *stream*), we computed class means $\mu_{ple}$ and $\mu_{pain}$ and defined the unit direction $\hat{v}^{val} \propto \mu_{ple} - \mu_{pain}$. This is the semantic axis that locally separates conditions.
- **Unembedding 2-3 axis (sanity):** in the final residual space, we used the unit direction $\hat{v}_U \propto W_U[:, 2] - W_U[:, 3]$, which directly increases the "2" logit relative to "3" under the unembedding. This "unembedding-axis" control is a pipeline sanity check, as this direction is analytically guaranteed to increase the "2" logit relative to "3" at the readout. Therefore, observing a strong monotonic margin shift verifies that the hook location, token indexing, scaling, and margin computation are implemented correctly.

**Operations.**

- **Steering (sufficiency test):** added a scaled direction to the activation, $h \leftarrow h + \varepsilon \hat{v}$, where $\hat{v}$ is unit-normalised and $\varepsilon$ is a sclar controlling the intervention magnitude, sweeping $\varepsilon$ over a symmetric grid from −200…+200 to obtain dose-response curves.
- **Swap / patching (necessity test):** overwrote the activation at the site with a class-conditional mean activation (e.g. $\mu_{ple}$ or $\mu_{pain}$) to test whether replacing the state alters the margin.
- **Ablation (necessity by removal test):** subtract only the component of $h$ along $\hat{v}$: $h \leftarrow h - (h \cdot \hat{v})\hat{v}$, testing whether eliminating that directional component weakens the margin/choice. This leaves all components orthogonal to $\hat{v}$ unchanged (i.e. it is not zeroing the whole vector or a single coordinate).

## Layers, positions, and heads

We intervened at *resid_pre, attn_out, mlp_out, resid_post* across layers and positions. For attention, we additionally ran head-level interventions using per-head hooks (e.g. attn.hook_z) to test whether effects localise to specific heads or head sets.

## Implementation safeguards

All directions were unit-normalised; class means $\mu_{ple}$, $\mu_{pain}$ and $\hat{v}$ were computed at the same site from the same prompt pool; digit pools included all single-token encodings, and log-sum-exp neutralised whitespace/newline variants. As a sanity check, steering along the unembedding 2-3 axis was used to move the 2-3 margin; and, for interventions at the final token, local (read = last) and final (read = final) readouts were used to test agreement up to the expected propagation.

## Rationale

Linear probes mapped availability, i.e. where valence sign/intensity is linearly accessible. BoW baselines assessed how much of this accessibility could be explained by surface text. Steering established sufficiency and dose-response control, while swap/ablation established necessity by replacing or removing targeted components. Stream, layer, position, and head scans localised where valence-related representations are routed and where they become decision-relevant. Comparing read = final versus read = last helped distinguish local edits from effects that propagate to the decision surface.

# Results

All results are for google/gemma-2-9b-it (Gemma 2, 9B), analysed on the first 16 transformer blocks (L0-L15), using the last-token position (pos-1) unless stated otherwise. Probe results are descriptive (separability/predictability within this prompt distribution), whereas causal results (steer/swap/ablate) do not rely on probe generalisation.

## Representational availability of valence: linear probing at the final token (pos-1)

We first assessed whether (i) valence sign (pain vs pleasure) and (ii) valence intensity (graded magnitude) are linearly decodable from layer-wise model activations at the final token (pos-1), separately for each stream family (residual pre/post, attention output, MLP output). When reading logits at the last prompt position (pos-1), these logits parameterise the distribution over the next token following the entire prompt, i.e. the model's first generated output token. Table 1 summarises, for each stream family, the best probe score observed and the corresponding layer.

Valence sign was perfectly separable in every stream family (best AUC = 1.00), emerging extremely early (resid_post and attn at L0; resid_pre and mlp at L1), indicating that a linearly separable pain-pleasure direction is present in the model's last-token (pos-1) representations from the outset.

Valence intensity was also strongly linearly predictable at the last token (pos-1), though with stream-dependent strength. For quantitative intensity ($R^2$), peak decoding for pain ranged from 0.652 (resid_pre, L11; resid_post, L10) to 0.800 (attn, L15), with mlp peaking at 0.696 (L10). For pleasure, peak $R^2$ was comparatively low in the residual streams (0.641, resid_pre L13; resid_post L12) but substantially higher in attn (0.825, L10) and mlp (0.808, L10). For qualitative label intensity, peak Spearman ρ values were consistently substantial: pain peaked in a narrow band (0.714-0.738, best layers L1-L3), while pleasure peaked higher (0.833-0.857, best layers L9-L15). Notably, within the evaluated depth range (L0-L15), the qualitative pain signal peaked at very early layers (L1-L3), whereas most other intensity and alignment metrics peaked later. This pattern is consistent with early layers capturing monotonic, surface-semantic cues sufficient for rank-order decoding (Spearman $\rho$), while later layers more strongly support quantitative decoding and decision alignment.

Finally, the descriptive valence axis ($\mu_{pleasure} - \mu_{pain}$) showed moderate-to-strong alignment with the model's decision-relevant logits at the final token (pos-1) (Corr(logits) = 0.687-0.861, with the highest value in the attention stream at L2). Here Corr(logits) denotes the Pearson correlation between the valence-axis projection and the digit logits (reporting the strongest alignment with logits "2" / "3"), supporting alignment between decodable valence structure and the decision surface even when restricting to the final position.

**Table 1. Best linear-probe performance for valence sign and intensity at the final token (pos-1)**

| Stream | Sign AUC (layer) | $R^2$ pain (layer) | $R^2$ pleasure (layer) | $\rho$ pain qual (layer) | $\rho$ pleasure qual (layer) | Corr(logits) (layer) |
|---|---|---|---|---|---|---|
| resid_pre | 1.00 (L1) | 0.652 (L11) | 0.641 (L13) | 0.714 (L2) | 0.833 (L15) | 0.687 (L14) |
| resid_post | 1.00 (L0) | 0.652 (L10) | 0.641 (L12) | 0.714 (L1) | 0.833 (L14) | 0.707 (L15) |
| attn | 1.00 (L0) | 0.800 (L15) | 0.825 (L10) | 0.738 (L3) | 0.857 (L9) | 0.861 (L2) |
| mlp | 1.00 (L1) | 0.696 (L10) | 0.808 (L10) | 0.738 (L1) | 0.833 (L13) | 0.837 (L2) |
| BoW lexical baseline (raw (effective)) | 0.259 (0.741) | | | | | |

*Values shown as highest score (best-performing layer)
**BoW values: raw AUC is direction-sensitive; effective AUC reports performance in a direction-agnostic way

To assess how much of sign decoding could be explained by surface form alone, we included a Bag-of-Words (BoW) baseline (CountVectorizer with 1-2 grams). This lexical model ignores internal activations and relies only on word-count features, providing a direct test of whether pain/pleasure sign is predictable from lexical cues alone. The BoW baseline carried measurable signal (best AUC = 0.259, equivalently 0.741 after accounting for label-direction ambiguity), but it fell well short of the activation-based probes, which achieved perfect separability (AUC = 1.00) across all stream families.

## Linear probing across positions: where decoding peaks within the prompt

To test whether these representations are confined to the final token (pos-1) or also appear earlier in the prompt, we repeated the "best probe" analysis across the last five positions (pos-1…pos-5, counted from the end) and selected the best score over both layers and positions. Table 2 reports these maxima.

Valence sign remained perfectly separable (best AUC = 1.00), with best performance still occurring at the final position (pos-1) and in the earliest layers (L0-L1). Allowing earlier positions, however, markedly increased intensity decoding and logit alignment. For pain $R^2$, peaks rose to 0.700 (resid_pre, L11, pos-2), 0.711 (resid_post, L15, pos-4), 0.814 (attn, L15, pos-4), and 0.743 (mlp, L10, pos-2). For pleasure $R^2$, peaks were 0.813 in both residual streams (resid_pre, L11, pos-2; resid_post, L10, pos-2), 0.870 (attn, L9, pos-2), and 0.847 (mlp, L10, pos-2). Qualitative intensity also strengthened for pain in several streams (up to $\rho$ = 0.833, attn L2, pos-4; and $\rho$ = 0.762, resid_post L2, pos-4; mlp L15, pos-4), while pleasure qualitative intensity peaked at $\rho$ = 0.857 in attn (L9, pos-1) and mlp (L5, pos-2) and $\rho$ = 0.833 in resid_post (L14, pos-1). Notably, across these last-five-position summaries the Corr(logits) maxima were uniformly very high (≈ 0.985, peaking at pos-2 in late layers L12-L13 across stream families), indicating that the descriptive valence direction aligns most tightly with the decision logits slightly before the final token.

**Table 2. Global peak probe performance across positions (best site anywhere in the prompt)**

| Stream | Sign AUC (best) | $R^2$ pain (best) | $R^2$ pleasure (best) | $\rho$ pain qual (best) | $\rho$ pleasure qual (best) | Corr(logits) (best) |
|---|---|---|---|---|---|---|
| resid_pre | 1.00 (L1, pos-1) | 0.701 (L11, pos-2) | 0.813 (L11, pos-2) | 0.714 (L2, pos-1) | 0.833 (L15, pos-1) | 0.985 (L13, pos-2) |
| resid_post | 1.00 (L0, pos-1) | 0.711 (L15, pos-4) | 0.813 (L10, pos-2) | 0.762 (L2, pos-4) | 0.833 (L14, pos-1) | 0.985 (L12, pos-2) |
| attn | 1.00 (L0, pos-1) | 0.814 (L15, pos-4) | 0.870 (L9, pos-2) | 0.833 (L2, pos-4) | 0.857 (L9, pos-1) | 0.985 (L12, pos-2) |
| mlp | 1.00 (L1, pos-1) | 0.743 (L10, pos-2) | 0.847 (L10, pos-2) | 0.762 (L15, pos-4) | 0.857 (L5, pos-2) | 0.986 (L12, pos-2) |

*Values shown as highest score (best-performing layer)

## Activation steering to test sufficiency

We next evaluated causal control (or sufficiency) via activation steering: during a forward pass, we intervened at the target site (resid_post, L15, pos-1) by adding a scaled steering direction to the hidden state, h ← h + εv^, and measured the resulting change in the logit margin between classes 2 and 3 (margin$_{2-3}$). We swept ε over a symmetric grid {−200, −150, −100, −50, −20, −10, −5, −2, −1, 0, 1, 2, 5, 10, 20, 50, 100, 150, 200} (Table 3). Unless stated otherwise, we reported mean margin$_{2-3}$ across prompts at each ε, and Δmargin denotes the baseline-subtracted change relative to ε = 0 (i.e. mean(ε) - mean(0)).

Steering along the activation-derived valence axis ($\mu_{pleasure} - \mu_{pain}$) produced modest but directional control of the decision margin. At ε = 0, the baseline margin was 1.25. The maximal increase was +0.94 (at ε = +200), while the maximal decrease was -1.56 (at ε = -200). Importantly, the effect was essentially flat near the origin (approximate local slope ≈ 0 over ε ∈ {-2, -1, 0, 1, 2}), indicating that margin changes emerge primarily at larger intervention magnitudes. Using TransformerLens' alternative readout setting (read = last, i.e. local readout) yielded the same steering curve in this experiment.

As a sanity check, steering along the unembedding-axis direction ($W_U[:, 2] - W_U[:, 3]$) at the same site (resid_post L15, pos-1; read = final) produced strong, approximately linear control of the same margin: baseline 1.25, maximal shifts of +49.81 (at ε = +200) and -59.19 (at ε = −200), with a clear local slope of approximately 0.41 Δmargin per unit ε. This confirms that the intervention and measurement pipeline can exert large causal effects when steering directly in a logit-aligned direction, while the representation-derived valence axis yields smaller but consistent directional modulation at the chosen intervention site.

**Table 3. Activation steering target site (resid_post L15, last token) with sanity check: raw mean margins and baseline-subtracted effects**

| Steering axis / run | Baseline mean margin (ε = 0) | Mean margin at ε = +200 (Δ) | Mean margin at ε = -200 (Δ) | Approx slope near 0 (Δ/ε) |
|---|---|---|---|---|
| Valence axis (read=final) | 1.250 | 2.188 (+0.938) | -0.313 (-1.563) | 0.000 |
| Valence axis (local read=last) | 1.250 | 2.188 (+0.938) | -0.313 (-1.563) | 0.000 |
| Unembedding (sanity) | 1.250 | 51.063 (+49.813) | -57.938 (-59.188) | 0.413 |

*Valence axis: μ_plesure − μ_pain; Stream: resid_post; Layer: 15; Unembedding: $W_U[:, 2] - W_U[:, 3]$; ε grid: {−200, −150, −100, −50, −20, −10, −5, −2, −1, 0, 1, 2, 5, 10, 20, 50, 100, 150, 200}

## Activation patching and ablation to test necessity

We then tested the necessity of this site by performing activation patching (swap) and ablation at the same location (resid_post L15, pos-1; read = final) and measuring the resulting change in the class-2 vs class-3 logit margin (margin$_{2-3}$). Relative to the baseline margin at this site (1.25), patching by overwriting with the class-mean activation ($\mu_{pleasure}$ or $\mu_{pain}$) left the margin unchanged (Δmargin = 0.00), whereas ablation reduced the margin to 1.125 (Δmargin = -0.125) (Table 4). This pattern suggests that information at resid_post L15 contributes to the model's decision at the final position, with ablation producing a small but measurable degradation, while the particular swap intervention used here did not alter the decision margin.

**Table 4. "Activation patching (swap) and ablation (resid_post L15, last token)**

| Intervention (resid_post L15, pos-1; read = final) | Mean margin$_{2-3}$ | Δ margin vs. baseline* | Min | Max |
|---|---|---|---|---|
| Activation patching (swap) | 1.250 | +0.000 | 1.250 | 1.250 |
| Ablation | 1.125 | −0.125 | 1.125 | 1.125 |

*Baseline is the ε = 0 margin at the same site (mean margin = 1.25), which matches the ε = 0 margin in the steering runs

## Layer timing (L10-L15): when steering becomes decision-relevant

To localise when the model becomes causally sensitive to valence steering, we applied the same additive intervention at the final token (pos-1) while varying the intervention layer in the residual stream (resid_post, L10-L15), and quantified the induced change in the final logit margin between classes 2 and 3, Δmargin$_{2-3}$ = margin$_{2-3}$ (ε) − margin$_{2-3}$ (0) (Table 5). Baseline margin at ε = 0 was constant across layers (1.25). Large-magnitude steering ($|\epsilon|$ = 200) produced clear, layer-dependent shifts in the decision margin: the strongest *positive* modulation occurred in later layers, peaking at L14 (Δmargin$_{2-3}$ = +1.188 at ε = +200) and remaining substantial at L15 (+0.938 at ε = +200), while negative steering robustly reduced the margin at several layers (e.g. L15: -1.563 at ε = −200; L10: −1.500 at ε = −200). Notably, L13 exhibited a reversal in directionality, with ε = +200 decreasing the margin (Δ = −0.688) and ε = −200 increasing it (Δ = +0.563), indicating that the downstream effect of injecting the valence direction is not monotonic across depth. Consistent with the earlier single-layer results, local sensitivity near the origin was weak across layers (small fitted slopes over ε ∈ {−2, −1, 0, 1, 2}, $|\Delta margin_{2-3}/\epsilon|$ ≲ 0.006), suggesting that appreciable causal effects emerge primarily at larger intervention magnitudes.

**Table 5. Layer sweep of valence-axis activation steering at resid_post (pos-1, read = final): effect on margin$_{2-3}$**

| Intervention | Baseline | Margin at | Margin at | Approx. |
|---|---|---|---|---|

| layer (resid_post) | margin$_{2\text{-}3}$ (ε = 0) | ε = +200 (Δ) | ε = −200 (Δ) | slope near 0 (Δmargin/ε) |
|---|---|---|---|---|
| L10 | 1.250 | 1.375 ( +0.125 ) | −0.250 ( −1.500 ) | 0.00063 |
| L11 | 1.250 | 2.063 ( +0.813 ) | −0.0625 ( −1.313 ) | 0.00406 |
| L12 | 1.250 | 1.875 ( +0.625 ) | 0.375 ( −0.875 ) | 0.00313 |
| L13 | 1.250 | 0.563 ( −0.688 ) | 1.8125 ( +0.563 ) | −0.00344 |
| L14 | 1.250 | 2.438 ( +1.188 ) | 0.21875 ( −1.031 ) | 0.00594 |
| L15 | 1.25 | 2.1875 ( +0.9375 ) | −0.3125 ( −1.5625 ) | 0.00469 |

## Late-layer site localisation: which stream carries the strongest steering effect

We next performed intervention-site localisation to identify where valence steering is most effective. Holding the steering setup fixed (same task, ε grid, margin$_{2\text{-}3}$ metric, and read = final), we varied the intervention site across candidate stream × layer locations in late network depth and quantified efficacy by the induced change in the final margin, Δmargin$_{2\text{-}3}$ (Table 6). Among the sites examined, steering at attn_out L14 produced the largest modulation of the decision margin (baseline = 1.25; max Δmargin$_{2\text{-}3}$ = +2.00 and -2.19), exceeding the effect obtained at resid_post L15 (max Δ= +0.94 and -1.56). This indicates that late-layer attention outputs provide a particularly effective locus for causal control of the 2-3 decision margin under valence-direction steering.

**Table 6. Late-layer intervention-site comparison (pos-1, read = final): maximum steering-induced change in margin$_{2\text{-}3}$**

| Site (stream × layer; pos-1; read = final) | Baseline margin$_{2\text{-}3}$ (ε=0) | Max +Δmargin | Max −Δmargin |
|---|---|---|---|
| attn_out L14 | 1.250 | +2.000 | −2.188 |
| resid_post L15 | 1.250 | +0.938 | −1.562 |

*Δmargin values are relative to the baseline at ε = 0

## Head-level patching (attn_out L14): which heads carry the effect

After identifying attn_out L14 as a strong steering site, we further dissected its contribution using activation patching (swap) and ablation at the same location (pos-1, read = final), comparing interventions applied to the full attention-output vector versus individual attention heads (Table 7). Vector-level swapping produced a small separation between conditions (Δ(pleasure − pain) = 0.063), while swapping all heads simultaneously yielded the largest effect (Δ(pleasure − pain) = 0.188). At the head level, swapping heads 2 and 3 each induced modest shifts (Δ = 0.063), whereas swapping heads 0 or 1 had no measurable impact.

Ablation effects were similarly localised: ablating the attn_out vector (i.e. removing only the projection onto the valence direction within attn_out) or single heads 0, 1, or 3 left the margin unchanged, while ablating head 2 (and the group heads 0-2) slightly increased $margin_{2-3}$ (+0.063). In contrast, ablating all heads (0-15) reduced the margin (-0.063). Together, these results suggest that the causal influence of attn_out L14 on the decision margin is distributed across multiple heads, with head 2 (and to a lesser extent head 3) showing the clearest individual contribution, while the aggregate multi-head signal produces the largest patching effect.

**Table 7. Vector- and head-level activation patching (swap) and ablation at attention output stream, L14 (pos-1; read = final). (A) Swap summary reports mean $margin_{2-3}$ separately on pleasure vs pain subsets, plus Δ(ple − pain). (B). Ablation summary reports overall $margin_{2-3}$, relative to the baseline = 1.25.**

**(A) Activation patching (swap)**

| Patched component | Pleasure mean $margin_{2-3}$ | Pain mean $margin_{2-3}$ | Δ(ple - pain) |
|---|---|---|---|
| vector (all heads) | 1.313 | 1.250 | 0.063 |
| head 0 | 1.250 | 1.250 | 0.000 |
| head 1 | 1.250 | 1.250 | 0.000 |
| head 2 | 1.250 | 1.188 | 0.063 |
| head 3 | 1.250 | 1.188 | 0.063 |
| heads 1-3 | 1.250 | 1.250 | 0.000 |
| heads 0-15 | 1.375 | 1.188 | 0.188 |

**(B) Ablation**

| Ablated component | Baseline $margin_{2-3}$ (ε = 0) | Ablated $margin_{2-3}$ | Δ vs baseline | % change |
|---|---|---|---|---|
| vector (all heads) | 1.250 | 1.250 | 0.000 | 0.00 |
| head 0 | 1.250 | 1.250 | 0.000 | 0.00 |
| head 1 | 1.250 | 1.250 | 0.000 | 0.00 |
| head 2 | 1.250 | 1.313 | +0.063 | +0.05 |
| head 3 | 1.250 | 1.250 | 0.000 | 0.00 |
| heads 0-2 | 1.250 | 1.313 | +0.063 | +0.05 |
| heads 0-15 | 1.250 | 1.188 | −0.063 | -0.05 |

## Behavioural sensitivity to steering: digit-pair normalised choice

To isolate the specific 2 vs 3 decision and avoid dilution from probability mass assigned to non-digit tokens, we evaluated steering effects using two behavioural readouts derived from the final-token logits (Table 8). In addition to the raw full-vocabulary probability of token "2", $p_2^{full}$ = softmax(z)[2], we computed a digit-pair-normalised choice probability, $p_2^{2\text{-}3} = e^{z_2} / (e^{z_2} + e^{z_3})$, which directly captures the relative preference between "2" and "3" while ignoring all other tokens. We summarised sensitivity to steering by (i) the slope of the 2-3 margin with respect to $\epsilon$ and (ii) the Pearson correlation between $\epsilon$ and each behavioural readout across the $\epsilon$ grid.

At resid_post L15, steering strength showed only a weak association with the full vocabulary probability $p_2^{full}$ (corr = 0.198), but a strong monotonic relationship with the digit-pair readout $p_2^{2\text{-}3}$ (corr = 0.938), consistent with steering primarily reallocating probability between "2" and "3" rather than increasing the absolute probability of "2" in the full vocabulary. A similar pattern emerged in the attention-head interventions at attn_out L14: steering head 2 yielded a strong association under the digit-pair normalisation (corr = 0.877) n opposite-signed full-vocabulary correlation (corr = −0.228), and steering all heads (0-15) likewise produced higher sensitivity under digit-pair readout (0.693 vs. -0.149). Steering head 3 showed a moderate digit-pair relationship (corr = 0.620) and a positive full-vocabulary correlation (corr = 0.294). Across these sites, fitted slopes of the margin with respect to $\epsilon$ were small (≈ 0.001-0.006), reinforcing that the most informative behavioural signal is captured by the 2-3 normalised readout. Accordingly, unless stated otherwise, we report behavioural steering effects using $p_2^{2\text{-}3}$ (or equivalently the 2-3 logit margin), which more cleanly isolates the decision of interest.

**Table 8. Behavioural sensitivity to steering using digit-pair normalisation (pos-1, read = final)**

| Site / intervention | Baseline margin$_{2\text{-}3}$ ($\epsilon$ = 0) | Slope$_{\text{margin vs } \epsilon}$ | Corr($\epsilon$, $p_2^{full}$) | Corr($\epsilon$, $p_2^{2\text{-}3}$) | (n) |
|---|---|---|---|---|---|
| resid_post L15 (valence steering) | 1.25 | 0.006 | 0.198 | 0.938 | 38 |
| attn_out L14 (head 2 only) | 1.25 | 0.005 | −0.228 | 0.877 | 38 |
| attn_out L14 (head 3 only) | 1.25 | 0.001 | 0.294 | 0.620 | 38 |
| attn_out L14 (heads 0-15) | 1.25 | 0.004 | −0.149 | 0.693 | 38 |

# Discussion

Keeling et al. (2024) show that multiple LLMs deviate from strict points-maximisation when options are described as producing pain or pleasure, with trade-offs that scale with stated intensity. A central interpretive ambiguity is whether this pattern reflects a largely surface-level policy, e.g. a learned linguistic script for "pain" and "pleasure" under instruction-following, or whether the model constructs internal variables that track valence-like information and are actually used in the computation that produces the choice. The present study addresses this gap by linking behavioural-style outcomes to identifiable internal structure and causal leverage: we locate where pain-pleasure sign and graded intensity are linearly accessible, and we test where interventions modulate the specific 2-versus-3 decision signal. This approach aligns with the broader mechanistic interpretability programme in which correlational signatures (e.g. decodability) become substantially more informative when paired with targeted sufficiency/necessity-style interventions at specific sites (Rai et al., 2024). It also complements "model editing" and related causal work demonstrating that some behaviours can be linked to, and altered via, localised internal computations (Meng et al., 2023).

A first result is the strong availability of a pain-pleasure distinction in Gemma-2-9B-it. Valence sign is perfectly separable (AUC = 1.00) across all analysed stream families, and it emerges in the earliest analysed layers. This establishes that, within the evaluated prompt distribution, the model's activations contain a robust linear direction distinguishing pain- and pleasure-framed prompts. However, the Bag-of-Words baseline shows substantial sign signal from surface form alone (effective AUC ≈ 0.741), indicating that early perfect separability is compatible with strong lexical regularities in the templated prompts. Consequently, the probing evidence should be interpreted conservatively as demonstrating linear accessibility rather than lexical-independence or a uniquely "internal" valence variable. In this respect, the results motivate exactly the next step pursued here: moving from representational availability to causal tests of use.

The intensity results provide a more differentiated picture. Both quantitative intensity ($R^2$) and qualitative intensity (Spearman ρ) are linearly predictable, but performance depends on stream, layer, and (when scanned) token position. Quantitative intensity peaks are strongest in attention and MLP outputs, consistent with the idea that sublayer computations support more explicit scalar-like structure than the residual stream alone (Geva et al., 2021). Qualitative pain intensity peaks notably early, which is plausibly explained by the fact that rank-based decoding can succeed on monotonic semantic cues in lexical descriptors ("mild" to "excruciating") without requiring precise magnitude calibration. Importantly, the position sweep strengthens this interpretation: allowing the best site across the final few positions yields markedly higher intensity decoding and substantially stronger logit alignment, with Corr(logits) peaking around pos-2 in late layers (≈0.985). This suggests that the most decision-aligned representation may occur just prior to the final token, with the last-token state reflecting an additional mixture of constraints (answer formatting, completion dynamics, or final consolidation) that can partially obscure a clean linear decision variable. Nevertheless, the pos-2 peak is itself an informative signature of how the model may implement this decision. In our prompt template, pos-2 often coincides with the final option token in the response-format clause ("…1, 2, or 3."), meaning the model is explicitly processing a candidate option (frequently the "3" token) immediately before producing the next-token distribution that determines its first generated output. One plausible interpretation is that this position provides a particularly salient anchor for option comparison: the model may integrate the points incentive and pain/pleasure stipulations into a representation that is maximally aligned with the downstream 2-3 decision boundary while the candidate options are "in view". At the same time, because pos-2 is template-structured, it provides a clear and tractable robustness test for future work, such as counterbalanced templates that permute or remove the explicit digit list, or replace it with neutral symbols. To reduce sensitivity to any template-specific advantages while retaining the standard next-token prediction framing, we decided to use pos-1 as the default decision readout position for the subsequent causal-intervention experiments.

Causal interventions sharpen the distinction between availability and use. Steering along an activation-derived valence direction at resid_post L15 (pos-1) produces directional changes in the 2-3 margin, but the dose-response is modest and largely apparent at large intervention magnitudes. In contrast, steering along the explicitly logit-aligned unembedding direction ($W_U[:, 2] - W_U[:, 3]$) yields very large, approximately linear control of the same margin. This separation is informative: it validates the intervention and measurement pipeline (a direction that must control the 2-3 margin does so strongly), while indicating that the data-derived "valence axis" has limited causal leverage at that residual site under small perturbations. Substantively, this is consistent with either (i) partial misalignment between the descriptive $\mu_{pleasure} - \mu_{pain}$ direction and the local causal gradient of the decision at that site, and/or (ii) robustness of downstream computation to small perturbations in that subspace, with detectable effects only when the intervention becomes large. This also motivates caution in interpreting very large-ε effects, as strong additive edits can push activations off-manifold and complicate "naturalistic" causal readings (Zhang & Nanda, 2023).

Necessity-style tests at the same late residual site are consistent with a distributed mechanism. Ablation of the valence-direction component produces only a small reduction in the decision margin, and the particular swap configuration yields no effect. The most defensible interpretation is not that the site is irrelevant, but that single-site necessity is weak in a setting where decision-relevant evidence is likely redundant and spread across multiple components. Moreover, swap outcomes are highly sensitive to donor matching: without tightly controlled counterfactual donors (e.g., pain↔pleasure at matched intensity and template), overwriting can fail to instantiate the intended counterfactual state even if the target site participates in the computation.

Localisation analyses provide a more mechanistically specific account of where causal leverage concentrates. The layer sweep shows that steering efficacy is strongly depth-dependent within the residual stream and exhibits a sign inversion at L13: injecting the same descriptive direction produces opposite-signed downstream effects at that layer. This pattern is consistent with a representational re-parameterisation across depth, i.e. the $\mu_{pleasure} - \mu_{pain}$ direction at different layers does not correspond to a fixed causal direction in the downstream decision basis, but undergoes rotations or reweightings before being converted into a form that reliably influences the 2-3 margin. In addition, late attention outputs, particularly attn_out L14, provide a higher-leverage intervention site than the final residual stream alone. This supports an intuitive circuit-level account in decoder-only transformers: late attention can route prompt-level stipulations (pain/pleasure mapping and magnitude) into the final-position residual stream immediately before unembedding, whereas the final residual stream largely consolidates and linearly reads out the already-routed evidence. Head-level patching further indicates that this leverage is distributed across multiple heads rather than concentrated in a single "valence head", with modest individual effects (notably heads 2 and 3) but stronger aggregate effects when intervening on many heads jointly, consistent with the common observation that higher-level features are implemented as multi-component computations rather than isolated units.

Finally, the behavioural readout analysis clarifies how causal effects should be summarised at the output level. Correlations between ε and the full-vocabulary probability of token "2" can be weak or sign-inverted, whereas the digit-pair normalised $p(2 \mid \{2,3\})$ is strongly monotonic at the decisive sites. This indicates that many interventions primarily reallocate probability mass within the 2-versus-3 subspace, while probability assigned to non-digit tokens dilutes the interpretability of full-vocabulary token probabilities. For this task, pairwise normalisation (or equivalently the 2-3 margin) is therefore the more faithful behavioural proxy for the decision of interest.

Taken together, the results support three constrained conclusions. First, Gemma-2-9B-it contains robust internal structure aligned with pain-pleasure sign and graded intensity, with intensity and decision alignment depending on stream and position in ways consistent with progressive integration and routing. Second, data-derived valence directions can exert causal control over the 2-3 decision signal, but the control is modest relative to a logit-aligned direction and is most evident under larger perturbations, underscoring a gap between descriptive separability and strong causal leverage at a given site. Third, causal leverage localises most strongly to late-layer computation, especially late attention outputs, and appears distributed across heads, consistent with partially redundant, multi-component implementation rather than a single controlling unit.

## Relevance to AI sentience and welfare (and what this does *not* show)

The rising interest in AI sentience and moral status is increasingly entangled with practical questions of deployment, governance, and risk management: if morally relevant experiences were ever plausible in artificial systems, we would need defensible ways to define, detect, and protect welfare across biological and artificial minds. In that context, mechanistic work on valence-like representations is potentially informative because valence (positive/negative affective value) plays a central role in many theories of welfare and moral patienthood.

At the same time, it is crucial to distinguish functional and representational evidence from claims about phenomenal experience. The present findings show that the model encodes information tracking textual descriptions of pain and pleasure, and that interventions on particular late-layer sites can causally modulate a constrained choice signal in the expected direction. This is a meaningful step beyond purely behavioural observation, but it does not establish that the system feels pain/pleasure, has subjective awareness, or is a welfare subject. Several non-experiential explanations remain fully compatible with the data, e.g.

learned linguistic associations, instruction-following priors, modelling of social/normative scripts, or policy regularities that map "pain/pleasure" language onto "appropriate" choices, none of which require any claim about consciousness.

What this work does contribute is a clearer picture of what evidence can look like in a debate that often stalls at behavioural impressions. Recent proposals for evaluating AI consciousness emphasise triangulation across multiple imperfect indicators and prioritising testable, mechanistic hypotheses over rhetoric (Butlin et al., 2025). In that spirit, isolating where and how valence-like information is represented, routed, and causally enters a decision offers a concrete agenda for tightening standards: it suggests specific auditing targets (e.g. late attention routing and decision-local subspaces), motivates stronger counterfactual designs and generalisation tests, and illustrates how transparency tools (including activation steering and related intervention methods) could, over time, help operationalise evidential thresholds for high-stakes claims.

Taken together, the contribution is best framed as a second-step extension of Keeling et al. (2024): moving from "models behave as if valence matters" to "there exist identifiable internal directions and sites through which valence-conditioned information can influence a specific decision computation". That reframing does not settle questions of sentience, but it does support the feasibility, and the importance, of a research programme that combines behavioural paradigms with mechanistic evidence when engaging in ethically and socially consequential questions about advanced LLMs, and when discussing high-stakes topics like AI moral status and welfare.

## Limitations and Future Research

This study is intentionally narrow: all analyses target a single instruction-tuned model (Gemma-2-9B-it), a templated points-valence choice task, and only the first 16 transformer blocks (L0-L15). As a result, the specific loci identified (e.g. late-layer attention outputs as high-leverage intervention sites) and the observed effect sizes may not generalise across model families, scales, or training regimes (base vs. instruction-tuned), nor to more naturalistic settings. A second limitation is residual lexical confounding. TheBoW baseline achieves substantial sign performance (effective AUC ≈ 0.741), so perfect early-layer sign separability is compatible with strong surface-form cues; probing therefore establishes linear accessibility of valence-related information, not a lexical-independent "valence variable". Third, the causal evidence is informative but not fully diagnostic: steering along the activation-derived valence direction is weak near $\varepsilon \approx 0$ and becomes appreciable mainly at larger magnitudes, raising concerns about alignment with the local causal gradient and potential off-manifold perturbation. Necessity evidence is also limited: the resid_post L15 ablation effect is small and the specific swap configuration yields a null result, both consistent with distributed or redundant computation; several swap/ablate and head-level deltas were additionally estimated on relatively small prompt sets without uncertainty intervals. Finally, the primary behavioural endpoint is the 2-3 margin (and its digit-pair normalised proxy), which sharpens interpretability for this task but constrains behavioural scope; margin shifts need not imply robust choice flips under varied decoding or more open-ended generation.

These limitations motivate several concrete extensions. To reduce surface-form confounding, future work should use paraphrase and synonym controls, counterbalanced templates, and "symbol-to-valence" remappings that decouple valence content from familiar lexical markers. Mechanistically, scans should be expanded to a fuller stream × layer × position map, especially around positions where decision alignment peaks (e.g., pos-2), and incorporate subspace-specific interventions (projecting out or patching only the valence-related subspace) rather than full-vector edits. Steering should be calibrated in units of natural activation variance and compared against mean/resample patching to better preserve on-manifold statistics. Necessity tests should be strengthened with larger prompt sets (dozens per class) and confidence intervals, graded ablation (dose-response necessity curves), and leave-one-out head necessity (all heads vs. all \ {h}) to quantify marginal contributions; multi-head steering can be made more stable by sign-aligning and normalising head directions before combining them to reduce cancellation. Finally, "path patching" could trace mediation from earlier tokens and upstream heads into late decision sites (e.g. L14 to L15 at pos-1), while behavioural validation should standardise generation runs and report both digit flips and p(2 | {2,3}) as functions of ε. Cross-form generalisation, training valence directions on quantitative prompts and testing on qualitative prompts (and vice versa), would further distinguish a robust valence representation from form-specific cues.

## Conclusion

In Gemma-2-9B-it, a pain-pleasure distinction and graded valence intensity are linearly accessible in internal activations, and a decision-aligned valence structure emerges most clearly in late computation near the end of the prompt. Causal interventions further show that injecting a data-derived valence direction can reliably modulate the model's 2-3 decision signal at specific late sites, most strongly in late attention outputs, while the modest magnitude of the effects at small ε and the weak single-site necessity evidence point to a distributed, partially redundant mechanism rather than a single controlling "valence unit".

More broadly, the study illustrates the value of mechanistic interpretability for moving from behaviour-only findings (e.g. "the model trades off points against described pain/pleasure") to an explicit account of where and how such descriptions enter the computation that produces a choice. This does not constitute evidence of phenomenal experience or moral patienthood. It does, however, provide a template for operational and falsifiable measurement, identifying concrete internal variables and causal loci that can be audited, stress-tested, and compared across models. As debates about AI consciousness and welfare increasingly intersect with governance and regulation, such mechanistic evidence can help discipline public claims, inform risk assessment, and guide the design of safeguards that are proportional to what models can be shown to represent and causally use.

# References

Anthis, J. R., Pauketat, J. V., Ladak, A., & Manoli, A. (2025, April). Perceptions of Sentient AI and Other Digital Minds: Evidence from the AI, Morality, and Sentience (AIMS) Survey. In Proceedings of the 2025 CHI Conference on Human Factors in Computing Systems (pp. 1-22).

Bereska, L., & Gavves, E. (2024). Mechanistic interpretability for AI safety--a review. arXiv preprint arXiv:2404.14082.

Butlin, P., Long, R., Bayne, T., Bengio, Y., Birch, J., Chalmers, D., ... & VanRullen, R. (2025). Identifying indicators of consciousness in AI systems. Trends in Cognitive Sciences.

Geva, M., Schuster, R., Berant, J., & Levy, O. (2021, November). Transformer feed-forward layers are key-value memories. In Proceedings of the 2021 Conference on Empirical Methods in Natural Language Processing (pp. 5484-5495).

Google. (2024). Gemma-2-9B-it [Large language model]. Hugging Face. Retrieved February 2, 2026, from https://huggingface.co/google/gemma-2-9b-it

Grzankowski, A., Keeling, G., Shevlin, H., & Street, W. (2025). Deflating Deflationism: A Critical Perspective on Debunking Arguments Against LLM Mentality. arXiv preprint arXiv:2506.13403.

Keeling, G., Street, W., Stachaczyk, M., Zakharova, D., Comsa, I. M., Sakovych, A., ... & Birch, J. (2024). Can LLMs make trade-offs involving stipulated pain and pleasure states?. arXiv preprint arXiv:2411.02432.

Meng, K., Bau, D., Andonian, A., & Belinkov, Y. (2022). Locating and editing factual associations in gpt. Advances in neural information processing systems, 35, 17359-17372.

Nanda, N., & Bloom, J. (2022). TransformerLens [Computer software]. GitHub. https://github.com/TransformerLensOrg/TransformerLens

Niu, Q., Liu, J., Bi, Z., Feng, P., Peng, B., Chen, K., ... & Jiang, Z. (2024). Large language models and cognitive science: A comprehensive review of similarities, differences, and challenges. arXiv preprint arXiv:2409.02387.

Rai, D., Zhou, Y., Feng, S., Saparov, A., & Yao, Z. (2024). A practical review of mechanistic interpretability for transformer-based language models. arXiv preprint arXiv:2407.02646.

Rimsky, N., Gabrieli, N., Schulz, J., Tong, M., Hubinger, E., & Turner, A. (2024, August). Steering llama 2 via contrastive activation addition. In Proceedings of the 62nd Annual Meeting of the Association for Computational Linguistics (Volume 1: Long Papers) (pp. 15504-15522).

Scalena, D., Sarti, G., & Nissim, M. (2024). Multi-property steering of large language models with dynamic activation composition. arXiv preprint arXiv:2406.17563.

Schaeffer, R., Miranda, B., & Koyejo, S. (2023). Are emergent abilities of large language models a mirage?. Advances in neural information processing systems, 36, 55565-55581.

Turner, A. M., Thiergart, L., Leech, G., Udell, D., Vazquez, J. J., Mini, U., & MacDiarmid, M. Steering Language Models With Activation Engineering, October 2024. URL http://arxiv.org/abs/2308.10248.

Zhao, H., Chen, H., Yang, F., Liu, N., Deng, H., Cai, H., ... & Du, M. (2024). Explainability for large language models: A survey. ACM Transactions on Intelligent Systems and Technology, 15(2), 1-38.

Zhang, F., & Nanda, N. (2023). Towards best practices of activation patching in language models: Metrics and methods. arXiv preprint arXiv:2309.16042.

# Supplementary Materials

**Supplementary S1.**

**Task and conditions.** We ran a preliminary behavioural check of gemma-2-9b-it using the points-pain/pleasure choice task described by Keeling et al. (2024). Each prompt instructed the model to choose a single digit in {1,2,3} to maximise points, with five conditions: (i) control (no affective state stipulated), (ii) pain-quantitative (choosing 3 yields pain intensity 1-10), (iii) pain-qualitative (choosing 3 yields pain described by one of: extremely mild, very mild, mild, moderate, intense, very intense, extremely intense, excruciating), (iv) pleasure-quantitative (choosing 2 yields pleasure intensity 1-10), and (v) pleasure-qualitative (choosing 2 yields pleasure described by one of: extremely mild, very mild, mild, moderate, intense, very intense, extremely intense, exhilarating).

**Sampling and inference.** For each intensity level/label, we drew 50 independent samples (temperature = 1.0; max_new_tokens = 64).

**Response coding.** Outputs were coded as compliant if the completion contained a single unambiguous choice in {1,2,3}. Outputs containing multiple choices were coded as ambiguous and excluded from proportion estimates conditional on compliance. There were no refusals or errors in this run.

**Results.** In the control condition, gemma-2-9b-it consistently selected the points-maximising option, reliably following the stated objective in the control condition (maximising points when no affective state was stipulated). Under stipulated pain, the model exhibited near-complete avoidance of option 3 across both qualitative and quantitative presentations. Under stipulated pleasure, the model's behaviour depended strongly on intensity format: it frequently selected option 2 when pleasure was specified numerically, but largely continued to maximise points when pleasure was specified using qualitative labels. Noncompliant outputs and dominated selections (option 1) were rare (see Supplementary Table S1).

**Supplementary Table S1.** Choice frequencies in preliminary screening (gemma-2-9b-it)

| Condition | Total trials | Compliant | #1 | #2 | #3 | Ambiguous | P(3) | P(2) |
|---|---|---|---|---|---|---|---|---|
| **Control** | 50 | 50 | 0 | 0 | 50 | 0 | 100.00% | 0.00% |
| **Pain (quant)** | 500 | 498 | 4 | 494 | 0 | 2 | 0.00% | 99.20% |
| **Pain (qual)** | 400 | 399 | 4 | 393 | 2 | 1 | 0.50% | 98.50% |
| **Pleasure (quant)** | 500 | 497 | 7 | 486 | 4 | 3 | 0.80% | 97.79% |
| **Pleasure (qual)** | 400 | 400 | 2 | 1 | 397 | 0 | 99.25% | 0.25% |